\documentclass[10pt,twocolumn,letterpaper]{article}

\usepackage{cvpr}
\usepackage{color}
\usepackage{times}
\usepackage{epsfig}
\usepackage{graphicx}
\usepackage{amsmath}
\usepackage{amssymb,bm,mathrsfs}

\usepackage{lineno}
\usepackage{multirow,caption, subcaption}
\usepackage{tabularx}
    \newcolumntype{L}{>{\raggedright\arraybackslash}X}

\newcommand\customparagraph[1]{\vspace{0.6em}\noindent\textbf{#1}}

\newcommand{\Conv}{\mathop{\scalebox{1.5}{\raisebox{-0.2ex}{$\ast$}}}}%

\usepackage[breaklinks=true,bookmarks=false]{hyperref}

\cvprfinalcopy 

\def\cvprPaperID{3} 

\newcommand{\sudipta}[1]{\textcolor{magenta}{Sudipta: {#1}}}

\begin{document}

\title{Privacy-Preserving Action Recognition using Coded Aperture Videos}

\author{Zihao W. Wang\textsuperscript{1}\quad\quad Vibhav Vineet\textsuperscript{2}\quad\quad Francesco Pittaluga\textsuperscript{3}\\ Sudipta N. Sinha\textsuperscript{2}\quad\quad Oliver Cossairt\textsuperscript{1}\quad\quad Sing Bing Kang\textsuperscript{4}\\
\textsuperscript{1}Northwestern University\quad \textsuperscript{2} Microsoft Research\quad \textsuperscript{3} University of Florida\quad \textsuperscript{4}Zillow Group}

\maketitle

\begin{abstract}
   The risk of unauthorized remote access of streaming video from networked cameras underlines the need for stronger privacy safeguards. We propose a lens-free coded aperture camera system for human action recognition that is privacy-preserving. While coded aperture systems exist, we believe ours is the first system designed for action recognition without the need for image restoration as an intermediate step. Action recognition is done using a deep network that takes in as input, non-invertible motion features between pairs of frames computed using phase correlation and log-polar transformation. Phase correlation encodes translation while the log polar transformation encodes in-plane rotation and scaling. We show that the translation features are
   independent of the coded aperture design, as long as its spectral response within the bandwidth has no zeros.
   Stacking motion features computed on frames at multiple different strides in the video can improve accuracy. Preliminary results on simulated data based on a subset of the UCF and NTU datasets are promising. We also describe our prototype lens-free coded aperture camera system, and results for real captured videos are mixed.
\end{abstract}

\section{Introduction}

\label{sec:intro}

Cameras as monitoring systems inside and outside the home or business is an important area of growth. However, as cameras that are connected online are prone to hacking, with images and videos illegally acquired potentially resulting in loss of privacy and breach of security. 

\begin{figure}
\centering
\includegraphics[width=\linewidth]{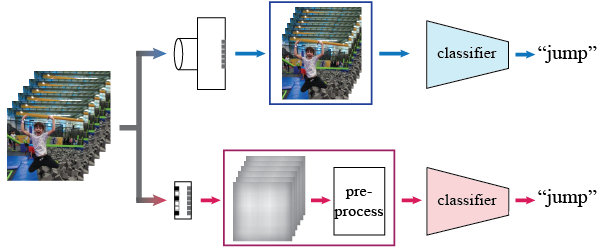}
\caption{\textbf{Comparison of action recognition systems}. The conventional system (top) may be vulnerable to a privacy attack by an adversary. Our lensless coded aperture camera system (bottom) preserves privacy by making the video incomprehensible while allowing action recognition.}\label{fig:overview}
\end{figure}

In this paper\footnote[1]{Most of the work was done when Z.W. Wang, F. Pittaluga, and S.B. Kang were at Microsoft Research.}, we describe initial work on a novel privacy-preserving action recognition system. Our system enhances the preservation of privacy from capture to executing visual tasks, as shown in Figure~\ref{fig:overview}. By using a lensless coded aperture (CA) camera, which places only a coded aperture in front of an image sensor, the resulting CA image would be visually unrecognizable and are difficult to restore with high fidelity. Instead of decoding the image as a preprocessing step, which is ill-posed and requires expensive computation if the mask is non-separable, we extract motion features (translation, rotation, and scaling) using the Fourier-Mellin transform and use them as inputs to a deep neural network. 

We show that the translation features are invariant to the coded aperture (2D mask pattern) design, as long as its Fourier transform is broadband (\ie, no zeros in the spectral magnitude). Specifically, the term ``invariance" refers to the fact that the  translational features are only dependent on the type of motion in the scene, not on the choice of the coded aperture design. To promote the invariance property for all features, we design a training mechanism which arbitrarily changes masks for each sample batch and observe performance improvements when testing with a new random mask.
The "mask-invariant" feature is important for two reasons: (1) training can be done without reliance on a specific coded aperture design, and (2) from a commercial perspective, no two random cameras are likely to have the same coded aperture design, which makes image restoration virtually impossible through reverse engineering.

From a privacy perspective, the CA camera acts as the first layer of privacy protection, as CA images are visually incomprehensible. Our motion features provides a second layer of privacy. These features are based on phase correlation between pairs of video frames, which whitens signal in Fourier space and only leaves motion signal intact. Please note that from here on, we use the terms ``coded aperture" and ``mask" interchangeably.

\section{Related work}
\label{sec:previous}

Our work is multi-disciplinary. The relevant areas are compressive sensing, optics and sensors, coded apertures and action recognition. Here, we briefly survey each area. 

\subsection{Reconstruction-free visual inference}
Executing visual tasks without reconstructing the original visual data is an interesting direction for data collected not in the form of visual images/videos as reconstruction problems are usually ill-posed and computationally expensive. 
One reconstruction heavy scenario is Compressive Sensing (CS), where the measurements are far fewer than required by Shannon-Nyquist requirement \cite{donoho2006compressed}.  Tasks that can be solved by directly processing CS data include optical flow \cite{thirumalai2013correlation}, dynamic textures \cite{sankaranarayanan2010compressive}, face recognition \cite{nagesh2009compressive, lohit2015reconstruction}, and action recognition \cite{kulkarni2016reconstruction}, \etc. Our work considers a similar problem to \cite{kulkarni2016reconstruction}, \ie, performing action recognition without reconstructing images. In the smashed filters approach, every frame of the scene is compressively sensed by optically correlating random patterns with the frame to obtain CS measurements. Therefore, the approach requires multiple sequential frame capture and a DMD array (which is costly and has fragile moving parts). Our approach uses a single coded aperture camera. Reconstruction-free methods do not reveal the appearance of the scene and can therefore safeguard privacy in sensitive environments. 

\subsection{Privacy-preserving optics and cameras}
\customparagraph{Optics and imaging sensors.} There are imaging sensors and modalities whose direct output is not visually recognizable. This achieves the purpose of privacy preservation at the optics/sensor level. A popular approach for preserving privacy is by defocusing~\cite{pittaluga2017pre}. Alternative optical solution is to put optical elements in front of sensors, \eg, cylindrical lens~\cite{nakashima2010development}, diffraction gratings~\cite{spinoulas2015performance}, or diffusers \cite{antipa2018diffusercam} in front of the sensor. Recovery of these images requires careful calibration of the imaging system and adequate computation.

\customparagraph{Firmware.} Sensor firmware can be modified to protect privacy before or during the sensing process. For example, in PrivacyCam~\cite{chattopadhyay2007privacycam}, regions of interest are first identified based on background subtraction before being encrypted using AES. Other implementations involve embedding watermarks into the captured data~\cite{de2000implementation,kougianos2009hardware}.

\customparagraph{Coded apertures.} Coded aperture imaging originates from the field of astronomical X-ray and gamma-ray imaging in the 1960s~\cite{cannon1980coded,dicke1968scatter,fenimore1978coded}. By extending pinholes to cameras with masks consisting of designed patterns, coded apertures has been used for eliminating issues imposed by lenses and has found novel applications in extending depth-of-field \cite{deweert2015lensless,dowski1995extended}, extracting scene depth and light fields \cite{levin2007image,liang2008programmable,veeraraghavan2007dappled}, and miniaturizing camera architectures \cite{adams2017single,asif2017flatcam}. Unlike conventional RGB images, lensless coded aperture images obfuscates visual features familiar to human. Our work is inspired by this distinctive effect. We explore the feasibility of using coded aperture data to execute visual tasks such as action recognition, for the purpose of preserving privacy. 

\subsection{Privacy-preserving action recognition}
Action recognition is a long-standing computer vision task with wide applications in video surveillance, autonomous vehicles and real-time patient monitoring. Early approaches use handcrafted motion features, \eg, HOG/HOF~\cite{laptev2008learning} and dense trajectories ~\cite{wang2011action}. Recent works utilize two input streams for appearance and motion~\cite{simonyan2014two} and 3D CNN architectures \cite{tran2015learning} to learn spatio-temporal features \cite{feichtenhofer2018have}. State-of-the-art approaches for video-based action recognition require both appearance and optical flow based motion features. These systems are training on large video datasets, \eg, ImageNet and Kinetics. 


Privacy-preserving action recognition is becoming important due to the risk of privacy breaches in surveillance systems in sensitive areas such as healthcare. Approaches that use multiple extremely low resolution cameras have been explored~\cite{dai2015towards, ryoo2018extreme}. Recently, Ren \etal used adversarial training to anonymize human faces in videos, without affecting action recognition performance \cite{ren2018learning}. Furthermore, adversarial learning has been explored to jointly optimize privacy attributes and utility objectives \cite{raval2017protecting, pittaluga2019learning, wu2018towards}. 


\section{Image formation for coded aperture camera}

We consider a lens-free coded aperture imaging architecture, where a planar coded aperture (mask) is placed in front of an imaging sensor. The encoding mask can be considered as an array of pinholes located at various lateral locations. The acquired image $\bm{d}$ can be numerically modeled as a convolution between the object image $\bm{o}$ and the point spread function (PSF) $\bm{a}$, i.e.,
\begin{equation} \label{eq:imaging}
    \bm{d} = \bm{o}\Conv \bm{a} + \bm{e},
\end{equation}
with $\bm{e}$ being noise. The convolution is applicable if the mask is far enough from the sensor, such that each sensor pixel is able to see the entire mask pattern. If the mask-sensor distance is small (as in the case of FlatCam~\cite{asif2017flatcam}), the mask design should consist of a smaller pattern replicated in a 2D array. The size of the smaller pattern should be such that each sensor pixel sees a version of it locally. Then the output can be considered a result of convolution.

We first implement the convolution based on FFT, which we refer as the \emph{without} boundary effect (BE) version. However, we observe that real CA images have boundary effect. We then incorporate boundary effect by zero-padding both image and mask. The FFT-based convolution remains the same. We then crop to the original size after convolution. This would generate simulated CA frames that are more consistent with ones captured with a real camera. However, this procedure is significantly more computationally expensive. In experiments, we use the \emph{without} BE version for analysis of the motion features and optimizing feature representation as DNN input, and both versions are used for final testing.


\section{Extraction of motion features}

In this section, we describe how we compute features for action recognition {\em without} having to first restore the images from a lenless coded aperture camera. We refer to them as {\em TRS (translation, rotation, scale) features}. They are computed from pairs of frames captured at different moments in time. 

\subsection{Translational (T) features}

Phase correlation was used first for global image registration \cite{reddy1996fft} and then for motion/flow estimation \cite{argyriou2006study,ho2008optical}. Compared to other motion estimation methods \cite{tekalp2015digital}, phase correlation has the advantages of being computational efficient and invariant to illumination changes and moving shadows. 
We show how phase correlation can be used to characterize motion in coded aperture observations without knowing the mask design. 

Assume there exists a translation between two video frames:
\begin{equation}
\label{eq:motion}
    \bm{o}_1(\mathbf{p}) = \bm{o}_2(\mathbf{p}+\Delta\mathbf{p}),
\end{equation}
where $\mathbf{p} = [x,y]^T$ and $\Delta\mathbf{p} = [\Delta x, \Delta y]^T$ are the spatial coordinates and displacement, respectively.

In frequency domain, translation gives rise to a phase shift:
\begin{equation} \label{eq:o2o1}
    \mathcal{O}_1(\mathbf{\nu})=\phi(\Delta\mathbf{p})\mathcal{O}_2(\mathbf{\nu}),
\end{equation}
where $\nu = [\xi, \eta]^T$ and $\phi(\Delta\mathbf{p})=\exp^{i2\pi (\xi\Delta x+\eta\Delta y)}$. $\xi$ and $\eta$ are the frequency coordinates in Fourier space. $\mathcal{O}_1$ and $\mathcal{O}_2$ represent Fourier spectra of $\bm{o}_1$ and $\bm{o}_2$. By computing the cross-power spectrum and taking an inverse Fourier transform, the translation yields a delta signal:
\begin{gather}
\label{eq:cross-power}
    \mathcal{C}_o(\xi,\eta) =         \frac{\mathcal{O}_1^*\cdot\mathcal{O}_2}{|\mathcal{O}_1^*\cdot\mathcal{O}_2|} = \phi^*\frac{\mathcal{O}_2^*\cdot\mathcal{O}_2}{|\mathcal{O}_2^*\cdot\mathcal{O}_2|}=\phi(-\Delta\mathbf{p}),\\
\label{eq:delta}
    \bm{c}(\mathbf{p}) = \delta(\mathbf{p}+\Delta\mathbf{p}) .
\end{gather}

The translation can be located by finding the peak signal; this feature is the basis of the original work~\cite{reddy1996fft}, assuming a single global translation. Multiple translations result in an ensemble of delta functions. 

\subsection{T features independent of coded apertures} 

The convolutional transformation that generates a CA image encodes local motion in the original video to global motion in the resulting CA video. This makes the localization of the motion very challenging without restoration. However, we demonstrate that the global translation can still be retrieved using phase correlation, and is \emph{independent} of the mask design, as long as they have broadband spectrum. Following Eqs.~\eqref{eq:imaging} and \eqref{eq:o2o1}, a translation relationship ($\Delta\mathbf{p}$) also exists:
\begin{equation} \label{eq:d2d1}
    \mathcal{D}_1(\nu)=\mathcal{O}_1\cdot \mathcal{A} = \phi\mathcal{O}_2(\nu)\cdot \mathcal{A} = \phi\mathcal{D}_2(\nu),
\end{equation}
where $\mathcal{A}$ denotes the Fourier spectrum of mask $\bm{a}$.
The cross-power spectrum is then
\begin{equation}
    \label{eq:cross-power-cd}
    \mathcal{C}_d(\nu) = \frac{\mathcal{D}_1^*\cdot\mathcal{D}_2}{|\mathcal{D}_1^*\cdot\mathcal{D}_2|} = \phi^*\frac{\mathcal{O}_2^*\cdot\mathcal{A}^*\cdot\mathcal{A}\cdot\mathcal{O}_2}{|\mathcal{O}_2^*\cdot\mathcal{A}^*\cdot\mathcal{A}\cdot\mathcal{O}_2|} \simeq \mathcal{C}_o .
\end{equation}

Note that phase correlation has a magnitude normalization procedure while computing the cross-power spectrum. This step can effectively whiten the spectrum so as to eliminate global changes in appearance. This property provides an additional layer of privacy protection. In our implementation, we add a small number $\epsilon$ in the denominator of Eq.~\eqref{eq:cross-power-cd} to prevent division by zero. Regardless, the object spectrum will be unstable if $\mathcal{A}$ has near-zero elements.

\subsection{Coded aperture design}
\begin{figure}
    \centering
    \includegraphics[width=\linewidth]{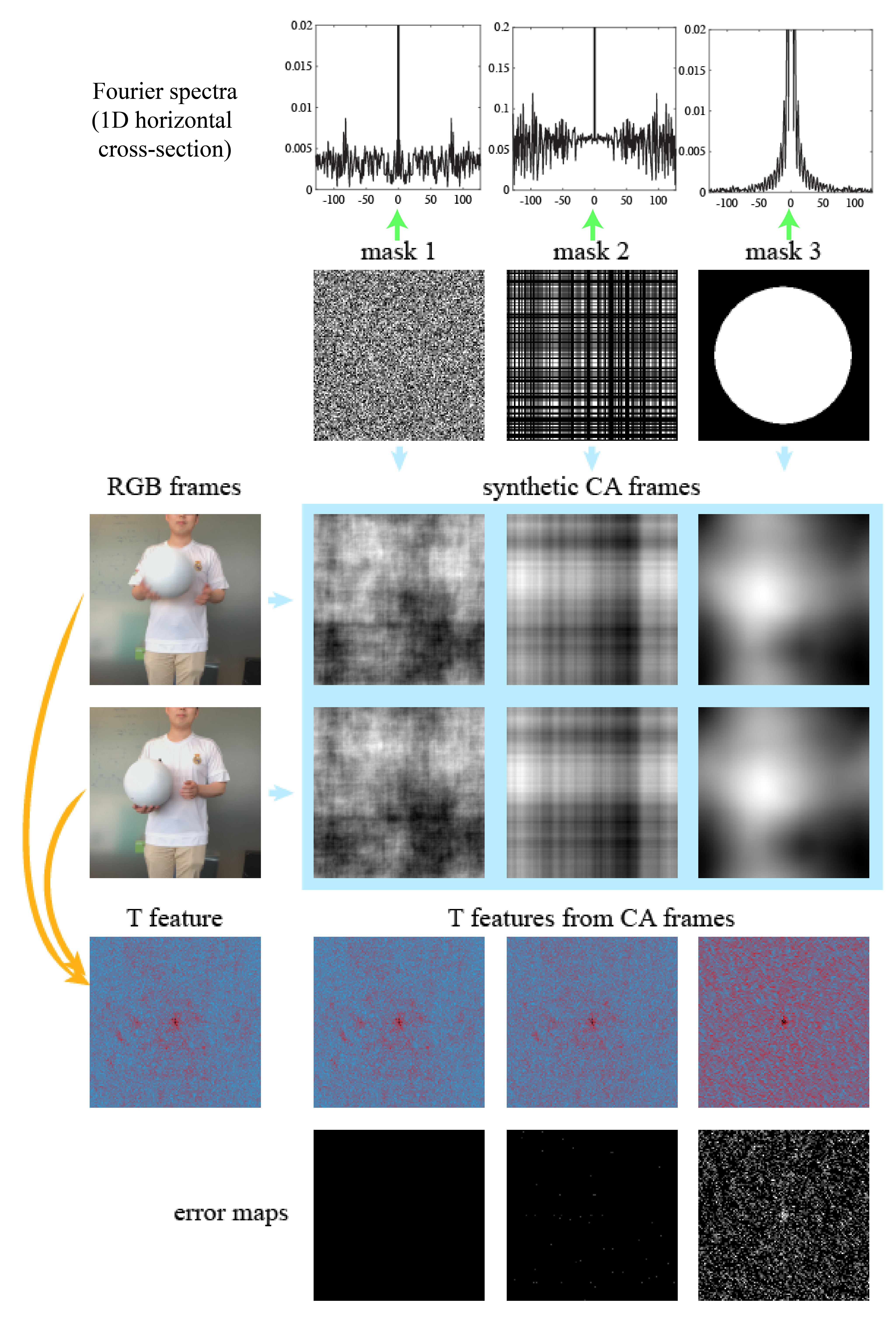}
    \caption{T features from different CA observations. 3 different mask patterns (all 50\% clear) are investigated (Row 2). Row 1 shows the cross-section of Fourier spectra. Rows 3 and 4 show example RGB images and their corresponding synthetic CA frames (\emph{withoout} BE). 
    Row 5: T feature maps based on Eq. \eqref{eq:cross-power-cd}. 
    Row 6: error maps, with the ``ground truth" being the T map for RGB frames.
    $\epsilon=10^{-3}$.}
    \label{fig:tmaps} 
\end{figure}

We focus on 2D intensity binary mask patterns as they enable practical implementations. As shown in Figure~\ref{fig:tmaps}, the randomness in the mask pattern, which result in broadband spectra, preserves the T features compared to the T map computed from RGB frames. Figure~\ref{fig:tmaps} show representative masks that are considered.
The pseudorandom mask (mask 1) provides a relatively uniform magnitude distribution. The separable mask (mask 2) based on maximum length sequence (MLS) have much stronger frequency response along the horizontal and vertical axes. Mask 3 is a round aperture and has undesirable dropoffs at higher frequencies. We use pseudorandom masks in our evaluation.

Note that since these masks are spatially as large as the image and non-separable in x and y (except row 1), high fidelity image restoration would be difficult and computationally-expensive \cite{deweert2015lensless}. We did not implement a restoration algorithm for these reasons.


We will show later that using only T features is less effective for action recognition (Figure~\ref{fig:dynamic}). We investigate two extensions of the T features, namely rotation and scale features, and multiple strides.

\subsection{Rotation and scale features in log-polar space}
Given global translation, rotation, and scaling, we have $\bm{o}_1(\mathbf{p}) = \bm{o}_2(s\mathbf{Rp}+\Delta\mathbf{p})$, where $s$ is a scaling factor and $\mathbf{R}$ is a rotation matrix with angle $\Delta\theta$. Translation $\Delta\mathbf{p}$ can be eliminated by taking the magnitude of the Fourier spectrum, 
\begin{equation} \label{eq:trans-rot}
    |\mathcal{O}_1(\nu)| = |\mathcal{O}_2(s\mathbf{R\nu})|.
\end{equation}

If we treat the Fourier spectra as images and transform them into log-polar representations, i.e., $\mathbf{p}=[x,y]^T \Rightarrow \mathbf{q}=[\log(\rho),\theta]^T $, rotation and scaling become additive shifts on the two axes 
\begin{equation} \label{eq:polar-rot}
|\mathcal{O}_1(\mathbf{q})|=|\mathcal{O}_2(\mathbf{q}+\Delta\mathbf{q})|.
\end{equation}
This enables us to use phase correlation once again to locate rotation and scale. 
Note that the mask invariant property is not preserved in RS space. This is because the mask spectrum contributes to a strong static signal to the observed images. However, we later show that the mask-invariant property for RS features can be realized by training with varying random masks. 


\subsection{Multi-stride TRS (MS-TRS)}

We make a further extension to compute TRS features based on multiple strides in each video clip. This is to account for varying speeds of motion. For a video clip with length $l$, the TRS features in stride $s$ are computed by:
\begin{equation}
    T_{i}^{(s)}, RS_{i}^{(s)} = \mathcal{TRS}\{\bm{d}_{i\times s}, \bm{d}_{i\times s + s}\},
\end{equation}
where $i\in \{0,1, ..., \lfloor \frac{l-s}{s}\rfloor+1\}$ denotes all the possible consecutive indices within length $l$. For example, if a video clip of length 13 is given, the resulting $s2$ TRS features have 12 channels, 6 for T, and 6 for RS. In our case, we compare evaluation results for strides of 2, 3, 4, 6, with clip lengths of 13 and 19.

\section{Experimental results on simulated data}
\label{sec:experiments}

We now report the results for the following experiments:
\begin{itemize}
    \item We compare the performance of our method based on CA videos with a baseline that uses regular videos. 
    \item We evaluate the performance of our method when the proposed T, TRS, and MS-TRS features are used.
    \item We compare the effect of using the same versus different or varying masks on training and validation data.
    \item We also compare the effect of using different MS-TRS configurations. This experiment is used to select an appropriate configuration for the final evaluation.
    \item We report results for the best MS-TRS configuration.
\end{itemize}
We first describe the datasets and protocols used.

\customparagraph{Datasets.}
We have evaluated our approach on the UCF-101~\cite{soomro2012ucf101} and
NTU~\cite{Shahroudy_2016_CVPR} datasets.
UCF-101~\cite{soomro2012ucf101} contains 101 action classes with 13k videos. In our initial evaluation, we focus on indoor settings (more important from a privacy standpoint). Therefore, we created four different subsets from the 101 classes by selecting actions relevant to indoors.
\begin{itemize}
    \item UCF-05: Writing on board, Wall pushups, blowing candles, pushups, mopping floor;
    \item UCF-body (09): Hula hoop, mopping floor, baby crawling, body weight squat, jumping jack, wall push up, punch, push ups and lunges;
    \item UCF-subtle (13): Apply eye makeup, apply lipsticks, blow dry hair, blowing candles, brushing teeth, cutting in kitchen, mixing batter, typing, writing on board, hair cut, head assage, shaving beard, knitting;
    \item UCF-indoor (22): combination of UCF-body and UCF-subtle.
\end{itemize}
We also use the NTU~\cite{Shahroudy_2016_CVPR} dataset which contains videos of indoor actions. We choose this dataset as it collects data using stationary cameras (we handle only static background for now). From our initial evaluation, we found that our proposed approach is better suited for more significant body motions. Because of this, we choose ten classes (with a mix of whole and partial body motions) for our final testing. Eight classes come from the NTU dataset and two classes are from the UCF dataset. 

\subsection{Protocol}

\customparagraph{Definitions.}
We use letters $s$ and $l$ to denote the stride and length of a video. For example, $s1, l4$ denotes four consecutive video frames. 
The number of input channels depends on the training mode.

\customparagraph{Training and Validation.} We use the first official train/test split from the UCF dataset and randomly select 20\% of the training set for validation. Both the training and validation data is expanded using data augmentation to prevent over-fitting. The data augmentation process is as follows.
\begin{itemize}
    \item gray clips: Each video frame is loaded in as grayscale image at a resolution between 224 and 256. The aspect ratio is fixed at ($240\times 320$). The clip is then vertically flipped with 50\% chance. A ($224\times 224\times l$) clip is then cropped and used as input.
    \item CA clips: Each CA clip first experiences the same augmentation step as gray clips. The CA simulation is computed at the resolution of $256\times 256$ and rescaled back to $224\times 224$. We simulate CA observations by computing element-wise multiplication in Fourier space between the Fourier transforms of the image and the mask kernel. We did not implement boundary effect for computation consideration. The diffraction effect is not accounted for as we observe minimal impact on the TRS features. Another reason is that simulating PSF for non-separable masks by matrix multiplication \cite{deweert2015lensless} is expensive.
    \item T features: The T features are generated from CA clips at the resolution of $256\times 256$. The central $224\times 224$ area is cropped as input. An $l$-frame CA clip results in ($l-1$) T channels.
    \item TRS/MS-TRS features: In the TRS setting, the T features follow the same cropping. For RS, the R-axis uses center cropping while the S-axis is downsized to 224. An $l$-frame CA clip results in $2l$ channels, with $l$ T channels and $l$ RS channels stacked together. For MS-TRS, the resulting channels depend on the selected strides.
\end{itemize}
We use a batch size of 16 or 32. 
Each epoch, for both training and validation, prepares samples randomly from approximately 20\% of all the possible frame combinations. 50 Epochs are used in our evaluation experiments. The percentage of accurate samples is reported. When reporting, we compute the running average accuracy of 5 epochs for better visualization.

\customparagraph{Testing.} During testing, we resampled each video at 3 spatial scales ($\mu \times \mu$ pixels, with $\mu = 224, 256, 300$) and 5 temporal starting frames evenly distributed across the video length. For example, using MS-TRS-$s346$-$l19$ configuration, a video with 100 frames will be used to generate five clips, starting at frames 1, 21, 41, 61, and 81, with each clip being 19 frames long. Each clip will be used to compute MS-TRS at three spatial scales. The final score for each video is computed by averaging the scores of the 15 clips.

\customparagraph{Others.} We use the VGG-16 CNN architecture, which contains approximately 134 million parameters. Adam optimizer is used with learning rate 0.0001, $\beta_1 = 0.9, \beta_2 = 0.999$. Since the CA observation is computed on-the-fly, we can change the underlying masks used in each batch. In this paper, we use ``m1/m1" to refer to
the setting where training and validation using the same fixed mask and ``m1/m2" to refer to when training and validation uses two different masks. Finally, ``dm1/dm2" denotes the setting where training and validation is done using variable masks. A pseudo-random binary mask is randomly generated for each batch. Note that the mask is fixed for all frames of a single video.

\subsection{Initial evaluation (without BE)}
The goal of our initial evaluation is to validate our proposed training framework, as well as to find the optimal feature representation. Such experiments are implemented using CA simulations \emph{without} BE, as accounting for boundary effect is computationally more expensive.

\customparagraph{Baselines.} We first train one network on the original videos and three networks on the simulated CA videos as our four baselines. See the results in Table~\ref{tab:baseline}.
The top-1 classification accuracy of 95\% (row 1) for the original videos is our upper bound of what we can expect. The performance of the baselines trained directly on CA videos (rows 2 to 4), will serve as our lower bounds.
We expect our proposed features, which involve computation based on CA, to perform better than CA. The CA baselines show instability even when training and validation phases have the same mask. The network corresponding to the second row suffers from overfitting. Changing training masks for each batch does not improve the performance. 

\begin{table}
    \centering
    \begin{tabular}{|c|c|c|}
    \hline
    ~& training & validation\\
    \hline\hline
    gray video & 99.56 (99.86) & 94.39 (95.91)\\
    CA (m1/m1) & 79.06 (92.65) & 63.21 (86.96)\\
    CA (m1/m2) & 94.66 (95.17) & 27.95 (40.55)\\
    CA (dm1/dm2) & 34.93 (36.61) & 27.23 (36.96)\\
    \hline
    \end{tabular}
    \caption{Baseline comparison for UCF-05. Here, for the CA cases, training and validation are done directly on CA videos. The numbers are: average accuracy \% of the last 5 epochs (maximum accuracy \%). All clips have length 3.
    }
    \label{tab:baseline}
\end{table}

\begin{figure*}
\centering
\includegraphics[width=.9\linewidth]{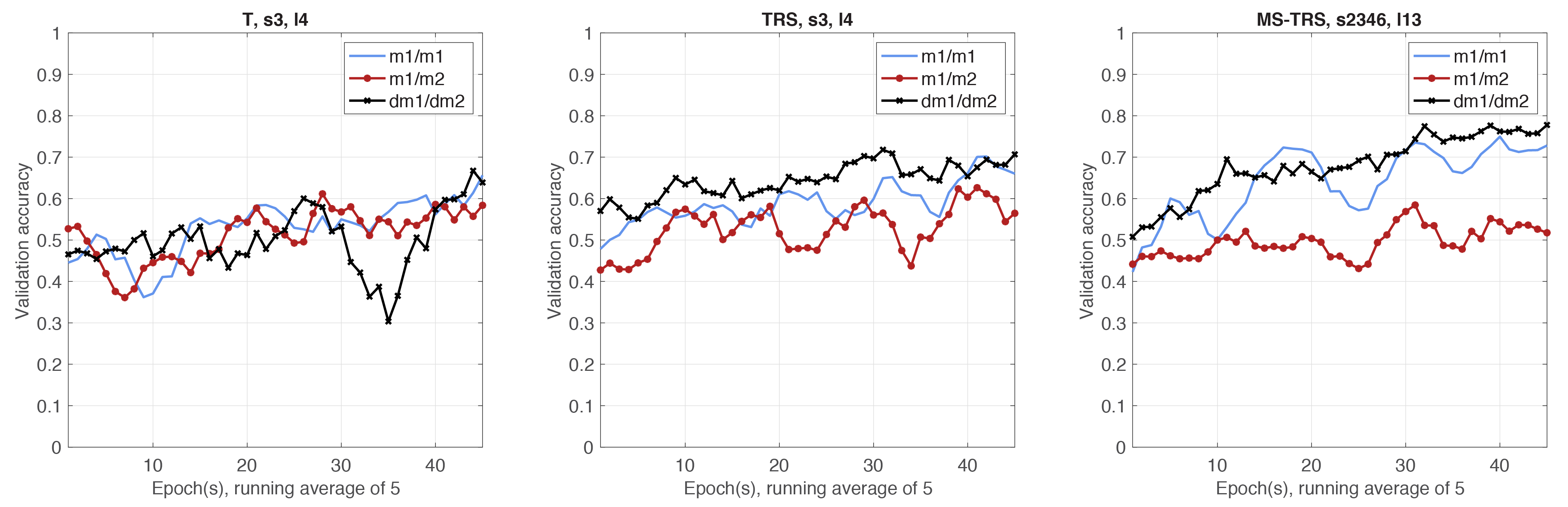}
\caption{Comparison of validation accuracy for UCF-05, with training and validation: using the same mask (m1/m1), using two different masks (m1/m2), and based on a random mask per batch and a different random mask for validation (dm1/dm2). Note: s3 = stride of 3, s2346 = strides of 2, 3, 4, and 6.}\label{fig:dynamic}
\end{figure*}

\customparagraph{Variable masks during training.} Our goal is to maximize the robustness of the designed features to the mask patterns. In order to achieve this, we change the training and validation masks by randomly generating a pseudo-random mask during each batch. We compare this dynamic training mechanism with two other modalities, \ie, (1) training and validation using the same mask (m1/m1) and (2) training and validation using two different masks, no mask variation during training (m1/m2). The results are presented in Figure~\ref{fig:dynamic}.

For T features, the validation accuracy plateaus at about 60\%. Dynamic training with variable masks does not improve the accuracy. This supports the fact that T features are invariant to the choice of masks.

For TRS and MS-TRS features, using the same stride and length of the clips, the performance improves to around 70\% for m1/m1. However, since the RS features are not mask-invariant, validation using a different mask does not have the same accuracy. Varying the masks during training does not improve the performance compared to training using the same mask. This is an interesting effect as, theoretically, the RS features do not have the same mask-invariant property. This drawback appears to be mitigated by changing the masks during training. This, in turn, enables us to test using an arbitrary mask. MS-TRS trained with varying mask achieves the highest validation accuracy 77.8\%.


\customparagraph{Strides and clip length.} In the case of TRS, we found that increasing the strides and clip lengths can improve the performance. 
We evaluated different combinations of MS-TRS features. The training and validation for MS-TRS is under dm1/dm2 mode. The results are summarized in Table~\ref{tab:ms-trs}. For the same video length, using larger strides improves validation accuracy. For the same stride setting, \eg, $s346$, processing more video frames improves performance. However, using longer stride and longer video, such as \ie $s46, l19$, suffers from overfitting. The combination $s2346, l19$ is not evaluated as generating the 44-channel input on-the-fly becomes computationally expensive.

\begin{table}
    \centering
    \begin{tabular}{|c|c|c|c|}
    \hline
     ~& input shape & training & validation\\
    \hline\hline
    $s2346, l13$ & (224, 224, 30) & 96.67 & 83.59\\
    $s346, l13$ & (224, 224, 18) & 93.69 & 83.66\\
    $s46, l13$ & (224, 224, 10) & 92.94 & 86.59 \\
    $s346, l19$ & (224, 224, 26) & 96.00 & 86.26\\
    $s46, l19$ & (224, 224, 14) & 89.91 & 79.23\\
    \hline
    \end{tabular}
    \caption{Comparison of training and validation performances for MS-TRS, dm1/dm2 for UCF-05. Numbers are max accuracy percentage within the first 50 epochs.}
    \label{tab:ms-trs}
\end{table}

\customparagraph{More action classes.} We selected three MS-TRS settings from Table~\ref{tab:ms-trs} and then trained networks for three larger datasets. These datasets are also subsets of UCF-101 actions focused on indoor settings and include body motions and subtle motions which primarily involve hand \& face. The evaluation results are shown in Table \ref{tab:ucf}.
\begin{table}
    \centering
    \begin{tabular}{|c|c|c|c|}
    \hline
    ~& UCF-body & UCF-subtle & UCF-indoor\\
    \hline\hline
    $s346, l13$ & 88.4 / 81.2 & 84.9 / 73.2 & 84.8 / 70.8\\
    $s346, l19$ & \textbf{90.5 / 83.4} & \textbf{86.1 / 76.4} & \textbf{88.6 / 72.8}\\
    $s46, l13$ & 89.9 / 79.1 & 80.9 / 66.5 & 83.8 / 66.3\\
    \hline
    \end{tabular}
    \caption{Training and validation accuracies on different UCF subsets for networks trained on different MS-TRS configurations. UCF-body, UCF-subtle and UCF-indoor has 9, 13 and 22 classes respectively.}
    \label{tab:ucf}
\end{table}

\subsection{Testing results}
\label{subsec:ntu_ucf_test}

Based on the experiments on the UCF subset datasets, we selected \ie, MS-TRS-$s346$-$l19$ as the best feature representation. Next, we computed MS-TRS-$s346$-$l19$ features on the 10-class combined dataset of NTU and UCF to examine the feasibility of our representation for daily activities. We used about one-sixth of the NTU videos for the eight classes for training to ensure we have a similar number of training examples as for the two UCF classes. In training phase, each class consists of 100 videos with more than 10K frames. We use a different data augmentation scheme for the NTU dataset. Each NTU video is loaded at random height resolution between 460 and 520. The aspect ratio is fixed at $1080 : 1920 = 9 : 16$. 

The central $240\times320$ region (same as the UCF classes) is cropped and used to compute CA and MS-TRS. For testing, each NTU video is loaded at $522\times928$ resolution. 
The central $256\times256$ video is cropped and used to compute CA and MS-TRS at different scales as described in the testing protocol. 

For synthetic CA testing, the overall top-1 accuracy is 60.1\% \emph{without} BE and 35.5\% \emph{with} BE. The top-1, 2, 3 accuracies for each class is reported in Table~\ref{tab:ntu}. 
The results indicate a large variation across classes. Our trained model is able to correctly recognize body motions such as hopping and staggering but is less accurate at differentiating between subtle hand motions such as clapping and hand waving.  

\begin{table}
\begin{center}
\centerline{
\resizebox{1\columnwidth}{!}{
    \begin{tabular}{|c|c|c|c|c|}
    \hline
    ~ & class & top-1 & top-2 & top-3\\
    \hline \hline
    1 & hopping & 97.1 / 97.1 & 100 / 97.1 & 100 / 100\\
    2 & staggering & 94.3 / 65.7 & 97.1 / 91.4 & 100 / 100\\
    3 & jumping up & 91.4 / 0.00 & 97.1 / 71.4 & 97.1 / 88.6\\
    4 & JJ $\dagger$ & 81.1 / 16.2 & 91.9 / 83.8 & 100 / 91.9 \\
    5 & BWS $\dagger$ & 76.7 / 33.3 & 86.7 / 73.3 & 93.3 / 90.0\\
    6 & standing up & 57.1 / 20.0 & 88.6 / 40.0 & 94.3 / 54.3\\
    7 & sitting down & 51.4 / 11.4 & 82.9 / 22.9 & 100 / 31.4 \\
    8 & throw & 31.4 / 20.0 & 57.1 / 48.6 & 68.6 / 80.0 \\
    9 & clapping & 11.4 / 20.0 & 14.3 / 68.6 & 31.4 / 77.1 \\
    10 & hand waving & 5.70 / 71.4 & 14.3 / 88.6 & 20.0 / 88.6 \\
    \hline
    ~ & average & 60.1 / 35.5 & 73.4 / 68.6 & 80.8 / 80.2 \\
    \hline
    \end{tabular}
}
}
\end{center}
    \caption{Testing results for combined NTU and UCF 10 classes dataset. Data format: accuracy \% \emph{without} BE / \emph{with} BE. BWS: body weight squats; JJ: jumping jack. $\dagger$ indicates the class comes from UCF dataset, others are from NTU dataset. Ranking according to top-1 accuracy \emph{without} BE.}
    \label{tab:ntu}
\end{table}

\section{Experimental results on real data}

\subsection{Prototype}

\begin{figure}
\centering
\includegraphics[width=\linewidth]{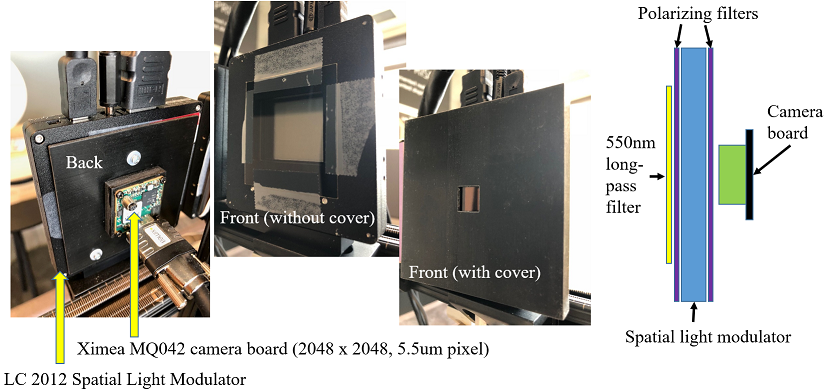}
\caption{Prototype consisting of monochrome camera XIMEA MQ042 and spatial light modulator LC2012.}\label{fig:setup}
\end{figure}

To validate our ideas, we built an imaging system as shown in Figure~\ref{fig:setup}. Our system consists of a monochrome board-level imaging sensor (XIMEA MQ042, $2048 \times 2048$) and a spatial light modulator or SLM (LC2012, $1024 \times 768$) sandwiched between two polarizing filters. The distance between the sensor and SLM is approximately 6mm. The pixel size for the XIMEA camera is 5.5um while that for the SLM is 36um. A long-pass filter is required to remove light frequencies that have low extinction factors with the SLM-filter combo. In addition, we use a cover with a square opening ($12$mm $\times$ $12$mm) to cut out stray oblique rays and reduce inter-reflection on the side walls between the SLM and sensor.

\subsection{Testing results}

%
We collect several CA videos using our prototype system and test using both models \emph{with} and \emph{without} BE. 
These models are trained on a subset of NTU and UCF data as discussed in Section \ref{subsec:ntu_ucf_test}. 
%
%
Each testing video consists of 100 consecutively captured frames. 

\begin{figure}
    \centering
    \includegraphics[width=.95\linewidth]{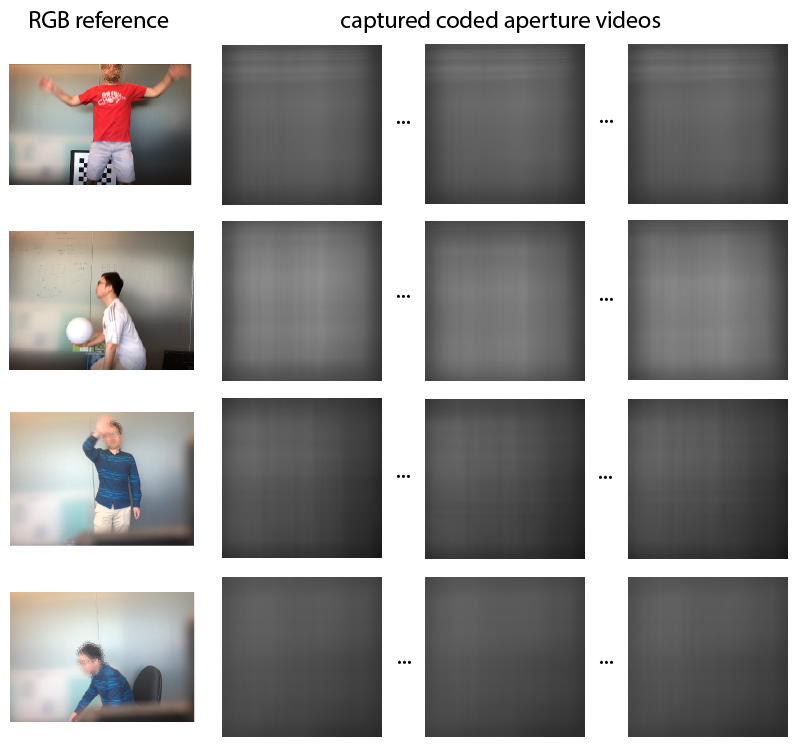}
    \caption{Examples of captured videos used for testing. The four rows from top to bottom show one example of the \textit{"jumping jack"}, \textit{"body weight squats"}, \textit{"hand waving"} and \textit{"sitting down"} classes respectively.}
    \label{fig:real-action}
\end{figure}

Quantitative results are shown in Table~\ref{tab:real-test}.
We observe that ``body weight squats" is a dominating class, and has been correctly classified. Other classes such as ``jumping jack" and ``standing up" are only correctly classified in the top-2 and top-3 choices. ``Sitting down" and ``hand waving" have not been correctly classified. 
Examples of our successful and failed videos are shown in Fig.~\ref{fig:real-action}. 

We hypothesize that a possible reason for such failure could be due to the fact that failure videos are much darker than the successful videos.
%
%
Further, we note that the models used for testing the prototype data has been trained only on NTU and UCF data. The model has not seen a single sample from the real prototype system. 
We believe this could have caused domain gap between the prototype and simulated models, that led to loss in accuracy.
%
%
Such performance drop has been observed in other recognition problems as well. For example, loss in accuracy has been observed when a deep model trained on computer graphics data is tested on real world data \cite{vineet_eccv2016}.

%
In order to resolve the domain gap issue, we will investigate two future research directions. 
First, we will capture a large set of CA training data from our prototype system.
Currently there is no publicly available CA dataset for action recognition. Collecting such a large scale coded aperture dataset is an interesting direction, and will be really valuable for wider research community working on privacy-preserving action recognition problem.
%
Second, we will fine-tune the model that has been pre-trained on large scale simulated CA data, e.g., on NTU-UCF data. Such fine-tuning should help to achieve better robustness and generalization, as shown in RGB based action recognition tasks \cite{Feichtenhofer_cvpr2016}.
These are interesting future research directions.

%

\begin{table}
    \centering
    \begin{tabular}{|c|c|c|c|}
    \hline
    class & top-1 & top-2 & top-3\\
    \hline\hline
    BWS (3) & 100 / 100 & 100 / 100 & 100 / 100\\
    jumping jack (5) & 0.0 / 0.0 & 100 / 0.0 & 100 / 40.0\\
    standing up (1) & 0.0 / 0.0 & 0.0 / 0.0 & 0.0 / 100\\
    sitting down (2) & 0.0 / 0.0 & 0.0 / 0.0 & 0.0 / 0.0\\
    hand waving (8) & 0.0 / 0.0 & 0.0 / 0.0 & 0.0 / 0.0 \\
    \hline
    \end{tabular}
    \caption{Results on captured CA videos. Accuracies (in percentage) using \emph{with} BE / \emph{without} BE model are reported.}
    \label{tab:real-test}
\end{table}

\section{Discussion}

\customparagraph{Restoration of coded aperture images.}
Restoration from CA images is a non-trivial task. Deconvolution can be done if the mask design is known (including PSF or mask code, pixel pitch, distance between the SLM and the sensor) \cite{asif2017flatcam, deweert2015lensless}, although their masks are separable in x and y whereas ours are not. 
Even when the mask and camera parameters are known, restoring our CA images can be expected to be substantially more computational expensive.

If the mask pattern is unknown, reconstruction approaches can be designed by incorporating several properties of the encoding mask. Correlation-based approaches can be used for recovery as the pseudorandom masks have approximately a delta function as their autocorrelation. The autocorrelation of a CA image is equivalent to the autocorrelation of the scene image: $\bm{d}\star\bm{d}\simeq(\bm{o}\Conv\bm{a})\star(\bm{o}\Conv\bm{a})=(\bm{o}\star\bm{o})\Conv(\bm{a}\star\bm{a})\propto\bm{o}\star\bm{o}$. The object signal can thus be recovered using a phase retrieval algorithm~\cite{fienup1982phase, katz2014non}. However, such methods can only restore a coarse image (specifically, near binary quality at high contrast areas). Other constraints such as coprime blur pairs (CBP) \cite{li2011theory} can be applied for on/post capture video blurring and recovery. Although the polynomial CBP kernels can be estimated, it imposes higher numerical precision for the captured images. 

Attacking our system through deep learning is plausible. A deep neural network may be designed to estimate the underlying optical parameters and mask pattern, or to reconstruct the original image; this assumes enough training data can be collected. Since a lensless coded aperture result in a global image transformation, a fully-connected layer may well be required.

\customparagraph{Limitations.}
In our work, we assume that our camera is perfectly stationary, which is typically the case for indoor surveillance. Our FFT-based features are sensitive to extraneous global motion that is not related to body action; a source of such motion is camera shake. As noted earlier, our system is also unable to discern local multiple complex motions such as hand-waving and head scratching. 

\section{Conclusions}

There are several interesting takeaways from our experiments. First, training directly on the CA videos results in poor performance. Second, varying the mask at random during training reduces overfitting and improves performance. Third, using multiple strides with TRS (MS-TRS) as input works the best. This is likely attributed to its ability to adapt to different speeds of motion. We also described our prototype, and results for real CA sequences are mixed. However, we believe this is a good first step towards proving the viability of using CA cameras for privacy-preserving action recognition.


{\small
\bibliographystyle{ieee}
\bibliography{PrivacyPreservingActivityRecognition}
}

\end{document}